\documentclass[a4paper]{spie}  %>>> use this instead for A4 paper
\makeatletter
\renewcommand{\@citess}[1]{\unskip~[#1]}

\makeatother

\usepackage{amsmath,amsfonts,amssymb}
\usepackage{graphicx}
\usepackage[colorlinks=true, allcolors=blue]{hyperref}
\usepackage{booktabs}

\title{On the Effectiveness of Adaptation Strategies for VLM-Based Federated Learning in Remote Sensing}

\author[a, b]{Simon Lösche}
\author[a, b]{Barış Büyüktaş}
\author[a, b]{Mathis Adler}
\author[c, d, e]{Angelos Zavras}
\author[c, d]{Ioannis Papoutsis}
\author[a, b]{Begüm Demir}

\affil[a]{BIFOLD - Berlin Institute for the Foundations of Learning and Data, 10587 Berlin, Germany}
\affil[b]{Technische Universität Berlin, 10623 Berlin, Germany}
\affil[c]{Orion Lab, School of Rural, Surveying and
Geoinformatics Engineering, National Technical University of Athens, 15772
Athens, Greece}
\affil[d]{Institute of Astronomy, Astrophysics, Space Applications
and Remote Sensing, National Observatory of Athens, 11810 Athens, Greece}
\affil[e]{Department of Informatics and Telematics, Harokopio University
of Athens, 17676 Athens, Greece}

\authorinfo{Send correspondence to simon.loesche@tu-berlin.de}

\begin{document} 

\maketitle
\begin{abstract}

Federated learning (FL) enables collaborative training of deep learning models across decentralized image archives without requiring data centralization. This paradigm is particularly relevant in remote sensing (RS), where legal regulations, privacy concerns, and bandwidth constraints restrict data sharing. However, the presence of training data heterogeneity across clients (known as non-IID data) can impede convergence and limit the generalization capability of the aggregated global model. To mitigate the adverse effects of training data heterogeneity, vision–language models (VLMs) can be leveraged in FL due to their transferable representations, which have demonstrated robustness under distribution shifts. However, their large parameter size may substantially increase communication overhead and local computational complexity in federated settings. Therefore, it is crucial to select an appropriate VLM adaptation strategy that balances the generalization ability with the communication and computational constraints. To address this issue, in this paper, we present the first comparative study of VLM adaptation strategies for FL in the context of RS image classification. We investigate full fine-tuning, encoder-specific fine-tuning, prompt learning, and low-rank adaptation (LoRA) tuning, and analyze them with respect to three criteria: 1) generalization capability under non-IID data, 2) communication overhead, and 3) local computational complexity. Experiments on BigEarthNet-S2, EuroSAT, RESISC45, and ImageNet reveal distinct trade-offs between task specialization, cross-domain generalization, and efficiency. Based on our findings, we derive a guideline for the selection of an appropriate VLM adaptation strategy in FL for RS image classification under different operational constraints. The code of this work is publicly available at \href{https://git.tu-berlin.de/rsim/FL-RS-VLM}{https://git.tu-berlin.de/rsim/FL-RS-VLM}.

\end{abstract}    
% Include a list of keywords after the abstract 

\section{Introduction}
\label{sec:intro}

Advances in remote sensing (RS) technologies have led to a rapid growth of RS images distributed across decentralized image archives (i.e., clients). In many operational scenarios, these archives cannot be centrally aggregated due to legal regulations, privacy concerns, commercial restrictions, or bandwidth limitations \cite{buyuktacs2024federated}. Consequently, conventional deep learning (DL) approaches that rely on centralized data access become impractical for a wide range of RS applications \cite{moreno2024federated}.

To enable collaborative model training under data access constraints, federated learning (FL) provides a decentralized training framework that allows multiple clients to jointly optimize deep neural networks without centralizing their local data. In FL, clients perform local model updates using their private data and share only model parameters with a central server for aggregation. This process makes FL particularly suitable for RS scenarios where data access is restricted \cite{buyuktacs2025multi}. However, applying FL to RS problems remains challenging due to the non-independent and identically distributed (non-IID) nature of training data across clients \cite{jiminez2024non}. In RS, data distributions on clients are inherently non-IID due to geographical, seasonal, and sensor-dependent variations, which result in distribution shifts and label imbalance across clients. As a result, locally optimized models may converge toward client-specific optima, which can slow down the federated optimization process and degrade the generalization capability of the aggregated global model \cite{xie2023fedkl}.

This issue can be addressed by leveraging vision–language models (VLMs) \cite{radford2021learning}, which learn transferable representations from large-scale image–text data and have demonstrated robustness to distribution shifts. This property makes them particularly relevant in federated settings, where non-IID data across clients can degrade convergence and generalization. Motivated by the robustness of VLMs under distribution shifts, recent studies in the ML and CV communities have investigated VLM-based methods within FL frameworks \cite{guo2023promptfl,pan2024federated,li2024global}. However, VLM-based methods have been seldom considered in the context of RS \cite{lin2025fedrsclip}. In \cite{lin2025fedrsclip}, a federated VLM framework is proposed (denoted as FedRSCLIP) for RS image classification. FedRSCLIP adopts prompt learning on top of a Contrastive Language-Image Pre-training (CLIP) model, where the backbone encoders are frozen and only a small set of prompt parameters is optimized in the FL setting. During training, clients update and transmit only the prompt parameters to the server for aggregation, which reduces communication overhead compared to full fine-tuning while mitigating the impact of non-IID data across clients.

Although VLM-based methods may improve generalization under non-IID data, their large parameter size substantially increases communication and computational complexity in federated settings. This raises the question of which VLM adaptation strategy is most suitable for FL in RS. Different adaptation strategies exhibit fundamentally different trade-offs with respect to their: 1) generalization capability under non-IID data; 2) communication overhead in terms of the volume of model updates exchanged between clients and the central server; and 3) computational overhead associated with local training on each client. Understanding these trade-offs is essential for the practical deployment of FL systems in RS, where clients may operate under different constraints.

To address this issue, in this paper we present the first comprehensive comparative analysis of VLM model adaptation strategies for FL in the context of RS image classification. To this end, we first provide an overview of the considered adaptation strategies and their integration into the FL framework. We then conduct both a theoretical comparison and an extensive experimental analysis of the considered strategies with respect to the three aforementioned criteria. Based on our analysis, we derive guidelines for selecting appropriate VLM adaptation strategies in FL for RS under different operational constraints.

\section{Related Work}
\label{sec:related_work}

In this section, we review existing studies that leverage VLMs in the CV and ML communities. Most existing works focus on parameter-efficient adaptation strategies to alleviate the communication cost caused by VLMs.

A prominent line of research investigates prompt learning within FL. In \cite{zhou_learning_2022}, context optimization (denoted as CoOp) is proposed as a parameter-efficient adaptation strategy in which both the image and text encoders are frozen and only a set of learnable context vectors is optimized. Extending this idea to the federated setting, in \cite{guo2023promptfl}, a federated prompt learning framework (denoted as PromptFL) is proposed, where soft prompts are optimized collaboratively across clients. The framework significantly reduces communication overhead by transmitting only prompt parameters instead of full model updates. In \cite{hou2025capt}, a federated prompt learning approach (denoted as CAPT) is introduced to address long-tailed and heterogeneous client distributions. CAPT employs both a global prompt and class-specific prompts, and clusters clients based on label distribution similarity to improve adaptation to rare classes while maintaining generalization. In \cite{pan2024federated}, a federated prompt portfolio framework (denoted as PromptFolio) is proposed, which combines a shared global prompt with client-specific local prompts. During training, both prompts are optimized locally, while only the global prompt is aggregated. Linear interpolation of global and local prompts enables a balance between generalization and personalization. In \cite{li2024global}, a federated prompt cooperation strategy (denoted as FedOTP) is introduced to align textual prompt features with visual representations. This strategy maintains a shared global prompt and personalized local prompts, using feature alignment to stabilize optimization under non-IID data. In \cite{cui2024harmonizing}, a federated prompt learning framework with generalization and low-rank personalization (denoted as FedPGP) is proposed to combine a shared global prompt with client-specific low-rank prompt adaptations. In addition, it introduces a contrastive objective to encourage the separation between global and personalized representations.

Beyond prompt-based methods, adapter- and low-rank-based strategies have also been investigated. In \cite{lu2023fedclip}, an adapter-based federated framework (denoted as FedCLIP) is proposed, where attention-based adapters are inserted into a frozen CLIP backbone. Only adapter parameters are updated and transmitted to reduce communication cost compared to full fine-tuning. In \cite{mitra2025fedvlm}, a federated VLM framework based on personalized low-rank adaptation (denoted as FedVLM) is introduced. This framework applies a personalized LoRA mechanism to an encoder–decoder VLM, where only part of the decomposed low-rank parameters are aggregated globally, while client-specific components remain local. This asymmetric aggregation strategy enables adaptation to heterogeneous client distributions while maintaining communication efficiency.

Other works explore alternative mechanisms for federated VLM adaptation. In \cite{wu2025faa}, a federated adversarial adaptation method (denoted as FAA-CLIP) is introduced to improve robustness under heterogeneous client distributions by employing a lightweight feature adaptation module combined with adversarial domain alignment. In \cite{shi2024clip}, a CLIP-guided FL framework (denoted as CLIP2FL) is proposed to guide client-side knowledge distillation by using CLIP as a teacher model. Instead of fine-tuning CLIP directly, a vision-only model is trained in the FL setting using multimodal supervision to mitigate long-tail bias.

%Although these works demonstrate different strategies for adapting VLMs in federated settings, their evaluation is primarily conducted on natural image benchmarks. Consequently, it remains unclear how different adaptation strategies behave under non-IID RS data.
\section{Methodology}

\subsection{Problem Formulation}

In the FL setting for RS, we consider a set of $K$ clients, denoted as 
$\{C_1, C_2, ..., C_K\}$. 
Each client $C_i$ has local training data $D_i$ consisting of $M_i$ labeled samples:

\begin{equation}
D_i = \left\{(\boldsymbol{x}_{i}^z, \boldsymbol{y}_{i}^z)\right\}_{z=1}^{M_i},
\end{equation}
where $\boldsymbol{x}_{i}^z$ represents the $z$-th RS image and 
$\boldsymbol{y}_{i}^z$ is its associated label. 
Depending on the task, $\boldsymbol{y}_{i}^z$ may correspond to single-label or multi-label annotations. 
The local data is private and not shared among clients in accordance with the privacy-preserving principles of FL. Each client $C_i$ maintains a local VLM, denoted as $\phi_i$, initialized with global model parameters. 
The VLM consists of an image encoder $f_{\theta_v}$ and a text encoder $g_{\theta_t}$, which map images and textual prompts into a shared embedding space. 
Given an input image $\boldsymbol{x}$ and a textual prompt $t_j$ corresponding to class $j$, the model computes normalized visual and textual representations as follows:
\begin{equation}
\boldsymbol{z}_v = \frac{f_{\theta_v}(\boldsymbol{x})}{\|f_{\theta_v}(\boldsymbol{x})\|_2}, 
\quad
\boldsymbol{z}_{t_j} = \frac{g_{\theta_t}(t_j)}{\|g_{\theta_t}(t_j)\|_2}.
\end{equation}
The class-wise logits are then obtained using scaled cosine similarity as follows:
\begin{equation}
s_j(\boldsymbol{x}) = \tau \, \boldsymbol{z}_v^\top \boldsymbol{z}_{t_j},
\end{equation}
where $\tau$ denotes a temperature parameter. The objective of client-side training is to minimize the empirical risk on the local data:

\begin{equation}
\label{eq:local_obj_vlm}
\begin{aligned}
\mathcal{O}_i (D_{i}; w_i)  
      &= \!\!\!\!\! \sum_{(\boldsymbol{x}_{i}^z, \boldsymbol{y}_{i}^z) \in D_i} \!\!\!\!\!
      \mathcal{L}(\phi_i(\boldsymbol{x}_{i}^z; w_i), \boldsymbol{y}_{i}^z),\\
     w_i^* &= \arg \min_{w_i} \mathcal{O}_i(D_i; w_i),
\end{aligned}
\end{equation}
where $\mathcal{L}$ denotes the task-specific loss function (e.g., binary cross-entropy for multi-label classification). 
The parameters $w_i$ include the trainable components of the VLM, depending on the selected adaptation strategy (e.g., full fine-tuning, encoder-specific tuning, prompt tuning, or low-rank adaptation). After local updates, the optimized parameters $w_i^*$ are transmitted to the central server, where they are aggregated to update the global model parameters $w$:

\begin{equation}
\label{eq:agg_vlm}
w = \sum_{i=1}^{K} \alpha_i w_i^*,
\end{equation}
where $\alpha_i$ denotes the aggregation weight for client $C_i$, which is typically proportional to $M_i$. Due to the large parameter size of VLMs, the iterative exchange of model updates in FL may incur substantial communication overhead and increased computational complexity at the client side. 
To mitigate these costs, different adaptation strategies restrict optimization to a subset of model parameters, denoted as $\theta_{\text{train}} \subseteq \{\theta_v, \theta_t\}$, while keeping the remaining parameters frozen. 
The choice of $\theta_{\text{train}}$ directly influences communication cost, local computational complexity, and generalization ability under non-IID data. In this work, we systematically analyze various VLM adaptation strategies within this federated optimization framework.

\subsection{VLM Adaptation Strategies}

%We investigate multiple CLIP-based adaptation strategies that differ in the choice of trainable parameters $\theta_{\text{train}} \subseteq \theta$. The considered strategies include full fine-tuning (FFT), partial fine-tuning, prompt learning, and low-rank adaptation (LoRA). This design choice determines not only the communication cost per round, as only $\theta_{\text{train}}$ is exchanged between clients and the server, but also the local computational complexity and the degree of representational flexibility under non-IID data. In this work, we focus on the fundamental VLM adaptation paradigms that differ in the subset of trainable parameters within a unified federated framework. While more advanced federated VLM approaches have been proposed, our objective is to provide a systematic comparison of the core adaptation mechanisms and their trade-offs in terms of generalization, local training complexity, and communication cost under non-IID RS data.

We investigate fundamental CLIP-based adaptation strategies that differ in the choice of trainable parameters $\theta_{\text{train}} \subseteq \theta$, including full fine-tuning (FFT), encoder-specific fine-tuning, prompt learning \cite{zhou_learning_2022}, and low-rank adaptation (LoRA) \cite{hu_lora_2022}. This design choice determines not only the communication cost per round, as only $\theta_{\text{train}}$ is exchanged between clients and the server, but also the local training complexity and the degree of representational flexibility under non-IID data. While more advanced federated VLM approaches exist, many build upon these fundamental adaptation strategies. We therefore provide a systematic comparison of the core adaptation strategies and their trade-offs in terms of generalization, local training complexity, and communication cost under non-IID RS data.

\paragraph{FFT.}

FFT updates both the image and text encoders of CLIP simultaneously (i.e., $\theta_{\text{train}} = \theta$). All model parameters are updated during training, allowing the full joint vision–language representation space to be optimized. This strategy provides the highest representational flexibility, since the full parameter space of CLIP can be optimized for the target task. As a result, it enables substantial realignment of visual and textual embeddings under domain shift. However, FFT incurs a high communication cost, since all model parameters must be transmitted during each communication round. In addition, local optimization of the full parameter space increases computational complexity on clients. In the presence of highly non-IID data, unconstrained local updates may result in client drift. Consequently, local models move toward client-specific optima, potentially slowing or destabilizing global convergence. Furthermore, full adaptation can increase the risk of catastrophic forgetting of pretrained representations, particularly when fine-tuning on domain-specific RS data.

\paragraph{Encoder-Specific Fine-Tuning.}

In encoder-specific fine-tuning, adaptation is restricted to a single modality while the other encoder remains frozen. 
In the image encoder variant, the text encoder is fixed and only the image encoder is updated (i.e., $\theta_{\text{train}} = \{\theta_v\}$). 
On the other hand, in the text encoder variant, the image encoder remains unchanged and only the text encoder is optimized (i.e., $\theta_{\text{train}} = \{\theta_t\}$). 
In both cases, only the selected encoder parameters are communicated in each federated round, reducing communication cost compared to FFT. The two variants differ in their adaptation behavior. Encoder-specific fine-tuning adjusts visual representations to client-specific image distributions while preserving pretrained language embeddings, which may be beneficial when distribution shifts mainly affect visual characteristics (e.g., geographical or sensor-dependent variations in RS images). 
Text encoder fine-tuning instead adapts the semantic embedding space to domain-specific label distributions while keeping visual features fixed. However, since VLM performance relies on cross-modal alignment, restricting optimization to a single encoder limits representational flexibility and may reduce robustness under highly non-IID data compared to FFT.

\paragraph{Prompt Learning.}

We employ prompt learning as a parameter-efficient adaptation strategy and consider CoOp \cite{zhou_learning_2022} as the primary prompt-based approach. In CoOp, only a set of learnable prompt embeddings is optimized. These prompts consist of a small sequence of context vectors in the CLIP embedding space that are prepended to the class names. For each class, the class name is inserted into the shared learnable prompt to generate class-specific textual features used for classification. Instead of modifying the internal representations of the encoders, CoOp adapts the textual conditioning of the model by reshaping the input prompts. This allows the model to better align pretrained representations with the target task while maintaining the original multimodal backbone. In FL, only the prompt embeddings are transmitted between clients and the server, resulting in minimal communication overhead and reduced local computational complexity. However, since the image and text encoders remain unchanged, adaptation capacity is limited to prompt-level adjustments, which may restrict flexibility under severe domain shifts or high non-IID data.

\paragraph{LoRA.}

LoRA \cite{hu_lora_2022} freezes the original CLIP parameters and introduces additional trainable low-rank matrices into selected transformer projection layers. Instead of directly updating the pretrained weights, LoRA restricts adaptation to a low-rank subspace while keeping the backbone unchanged. Only these inserted low-rank components are optimized and communicated during each communication round, significantly reducing communication overhead and local computational complexity compared to FFT. By constraining updates to a structured low-rank space, LoRA preserves pretrained multimodal representations while enabling efficient adaptation of internal transformer features. This can improve stability under non-IID data and mitigate catastrophic forgetting. However, the restricted update space may limit adaptation capacity when severe domain shifts are present.

\section{Dataset Description and Design of Experiments}

\subsection{Dataset Description}

Experiments were conducted on BigEarthNet-S2 v2.0 \cite{clasen_reben_2025}, EuroSAT \cite{helber_eurosat_2019}, RESISC45 \cite{cheng_remote_2017}, and ImageNet \cite{deng_imagenet_2009} datasets to evaluate the considered VLM adaptation strategies in FL. BigEarthNet-S2 was used as the federated fine-tuning dataset, while the remaining datasets were used exclusively for evaluation. BigEarthNet-S2 is a large-scale multi-label dataset derived from Sentinel-2 images and annotated using the CORINE Land Cover nomenclature. In our experiments, the official train–test split was followed and only the RGB bands of Sentinel-2 were used to ensure compatibility with the pretrained CLIP backbone. To simulate statistical heterogeneity across clients, images acquired over Austria, Belgium, Finland, Ireland, Lithuania, Serbia, Portugal, and Switzerland were partitioned by country such that each client received data from a single country. This resulted in non-IID data across clients due to regional variations in class frequencies and semantic characteristics. EuroSAT consists of 27,000 Sentinel-2 RGB images annotated with 10 classes. In our experiments, the official train–test split was followed. Due to its shared sensor modality and semantic proximity to BigEarthNet-S2, it enabled the assessment of cross-dataset generalization within the RS domain. RESISC45 contains 45 scene-level categories with substantial intra-class diversity and variations in spatial scale and viewpoint. It was used to evaluate generalization under increased scene diversity and distribution shift. We used a large-scale natural image dataset as a proxy to assess catastrophic forgetting of the pretrained representations after fine-tuning on the RS task. Specifically, we used the ILSVRC benchmark \cite{russakovsky_imagenet_2014}, a subset of ImageNet comprising 1,000 object classes. The dataset consists of approximately 1.2 million training images, 50,000 validation images, and 100,000 test images.

\subsection{Design of Experiments}

\paragraph{Modelling.}

All CLIP-based models were initialized with pretrained OpenAI CLIP ViT-L/14 weights. For CoOp, the maximum context length supported by the tokenizer given the class names was used, resulting in 61 learnable context vectors. LoRA was implemented using the HuggingFace PEFT library. Low-rank adaptation was applied exclusively to the query and value projection matrices within the transformer attention blocks, while all original pretrained weights remained frozen. The LoRA configuration used a rank of 8, a scaling factor of 16, and a dropout probability of 0.1. No bias parameters were optimized in either the adapters or the base model. Whenever additional parameters were introduced (e.g., in CoOp or LoRA), they were randomly initialized across all clients.

\paragraph{Prompt Construction.}

Class names were obtained directly from the respective datasets. Underscores were replaced with spaces prior to tokenization. For RS datasets, domain-specific prompt templates tailored to satellite images were adopted \cite{goyal_finetune_2023}. For ImageNet, natural-image prompt templates were used. All prompts were tokenized using the CLIP tokenizer. For each class, text features were generated by inserting the class name into the corresponding template. The resulting prompt embeddings were averaged per class to obtain a single text representation.

\paragraph{Training Protocol.}

For federated training on decentralized and non-shared RS image partitions, the same FL training procedure was adopted across all adaptation strategies. Model updates were aggregated using standard iterative model averaging after each communication round. Each client trained a local model initialized from the current global model parameters. In each communication round, local training was performed for three local epochs using a mini-batch size of 128. Optimization was carried out using AdamW. For CLIP-based models, the learning rate was set to $1 \times 10^{-5}$ with a weight decay of $1 \times 10^{-4}$. As a conventional convolutional neural network baseline, we employed a ResNet-50 architecture trained from scratch in the federated setting \cite{mcmahan2017communication}. For this baseline, a learning rate of $1 \times 10^{-4}$ without weight decay was used. All experiments were conducted on an NVIDIA A100, with clients trained sequentially within each communication round.

\paragraph{Metrics.}

We conducted experiments to compare the considered adaptation strategies in terms of classification performance, local training complexity, and communication cost. For multi-label classification (MLC) performance, we reported mean Average Precision (mAP). For single-label classification performance, we reported top-1 accuracy, top-5 accuracy, and recall. Local training complexity was measured in terms of floating-point operations (FLOPs) and multiply-accumulate operations (MACs). Communication cost was quantified by the number of trainable parameters exchanged between communication rounds.

\section{Experimental Results}
\label{cha:results}
In this section, we systematically compare the considered adaptation strategies in terms of in-distribution performance on BigEarthNet-S2, out-of-distribution performance on EuroSAT, RESISC45, and ImageNet. We further analyze the complexity of local training and the cost of communication. The main quantitative results are summarized in Table~\ref{results:performance1}.

\begin{table*}[htbp]
\renewcommand{\arraystretch}{1.2}
\caption{Performance comparison of the considered adaptation strategies. In-distribution performance on BigEarthNet-S2 is reported using micro- and macro-mAP. Out-of-distribution performance is evaluated on EuroSAT, RESISC45, and ImageNet using top-1 and top-5 accuracy as well as mean per-class recall. Bold and underlined values indicate the best and second-best results per column, respectively.}
\vspace{4mm}
\resizebox{\textwidth}{!}{
\begin{tabular}{@{}l cc ccc ccc ccc @{}}
\toprule
 & \multicolumn{2}{c}{\textbf{BigEarthNet-S2}} 
 & \multicolumn{3}{c}{\textbf{EuroSAT}} 
 & \multicolumn{3}{c}{\textbf{RESISC45}} 
 & \multicolumn{3}{c}{\textbf{ImageNet}} \\
\cmidrule(lr){2-3}
\cmidrule(lr){4-6}
\cmidrule(lr){7-9}
\cmidrule(lr){10-12}
\addlinespace[3pt]
\textbf{Adaptation Strategy} 
& micro-mAP & macro-mAP 
& top-1 & top-5 & recall 
& top-1 & top-5 & recall 
& top-1 & top-5 & recall \\
\midrule
Baseline & 50.5 & 35.7 & --- & --- & --- & --- & --- & --- & --- & --- & --- \\
\midrule
Zero-shot & 26.2 & 27.5 & \underline{44.8} & \underline{88.0} & \underline{44.8} & \textbf{69.2} & \textbf{95.3} & \textbf{69.3} & \textbf{75.6} & \textbf{94.6} & \textbf{75.6} \\
FFT & \underline{78.3} & \textbf{61.8} & 40.0 & 78.0 & 41.7 & 15.4 & 56.4 & 15.5 & 7.8 & 22.4 & 7.8 \\
Image Encoder Fine-tuning & 74.8 & 57.8 & 34.2 & 84.0 & 35.3 & 31.8 & 70.1 & 31.0 & 63.1 & 86.8 & 63.1 \\
Text Encoder Fine-tuning & 68.8 & 52.0 & \textbf{45.5} & \textbf{89.0} & \textbf{48.7} & \underline{54.9} & \underline{85.2} & \underline{54.8} & 61.9 & 86.5 & 61.9 \\
CoOp & 66.5 & 48.3 & 19.1 & 63.4 & 21.1 & 34.3 & 65.7 & 34.2 & 23.5 & 43.7 & 23.5 \\
LoRA & \textbf{79.1} & \underline{58.7} & 19.9 & 79.0 & 19.4 & 40.4 & 73.3 & 40.3 & \underline{75.1} & \underline{94.3} & \underline{75.1} \\
\bottomrule
\end{tabular}
}
\label{results:performance1}
\end{table*}

\subsection{In-Distribution Performance} 

All adaptation strategies substantially outperform zero-shot CLIP, confirming the benefit of task-specific adaptation in the federated setting. The strongest improvements are observed for LoRA and FFT. In terms of micro-mAP, LoRA improves performance by more than 52\% compared to zero-shot CLIP, while FFT achieves the largest gain in macro-mAP with an improvement of approximately 34\%. These results indicate that adaptation strategies that allow modifications across both image and text representations provide the most effective alignment to the RS task. Restricting fine-tuning to a single modality results in smaller gains. Image encoder fine-tuning improves micro-mAP by roughly 49\% over zero-shot CLIP, while text encoder fine-tuning yields an improvement of about 43\%. Although both approaches significantly outperform zero-shot CLIP, they remain below FFT and LoRA. This suggests that limiting updates to only one modality constrains the capacity of model to jointly adapt visual and semantic representations to the target task.
CoOp provides the smallest improvement among the adaptation strategies, increasing micro-mAP by approximately 40\% over zero-shot CLIP. Since CoOp modifies only learnable prompt tokens without updating backbone weights, its capacity to compensate for the substantial domain shift between natural-image pretraining and RS images is limited. In addition, all CLIP-based fine-tuning strategies achieve higher performance compared to the baseline. LoRA improves micro-mAP by almost 29\% over the baseline, demonstrating the advantage of adapting a large pretrained VLM in FL. These results highlight that leveraging pretrained multimodal representations provides a clear benefit over training a conventional CNN from scratch.

\subsection{Out-of-Distribution Performance}
\label{generalization}

As one can observe from Table~\ref{results:performance1}, the zero-shot model achieves the highest performance across most evaluation datasets. Zero-shot CLIP outperforms FFT by more than 29\% on ImageNet and 53\% on RESISC45. This indicates that fine-tuning on BigEarthNet-S2 leads to task specialization, which negatively affects generalization to other distributions. All adaptation strategies exhibit a noticeable performance degradation compared to zero-shot CLIP, with the exception of text encoder fine-tuning on EuroSAT, which improves top-1 accuracy by approximately 1\%. However, for RESISC45, text encoder fine-tuning remains around 14\% below zero-shot performance, although it performs substantially better than FFT and CoOp. The improvement on EuroSAT can be explained by the higher similarity between BigEarthNet-S2 and EuroSAT. In contrast, RESISC45 contains more diverse scene categories and exhibits a larger distribution shift. Image encoder fine-tuning results in reductions of more than 10\% on EuroSAT and 37\% on RESISC45 compared to zero-shot CLIP. As the distribution shift increases from EuroSAT to RESISC45, the performance gap widens further. This indicates that adapting only visual representations disturbs pretrained features that are beneficial for out-of-distribution performance, whereas adapting textual representations preserves visual features. On ImageNet, LoRA achieves 75.1\% top-1 accuracy, resulting in a reduction of less than 1\% compared to zero-shot CLIP and preserving most of the pretrained representation. In contrast, image encoder fine-tuning and text encoder fine-tuning achieve 63.1\% and 61.9\% top-1 accuracy, corresponding to reductions of approximately 13\% and 14\%, respectively. CoOp reaches 23.5\% top-1 accuracy, exhibiting a degradation of more than 52\%. The strongest forgetting is observed for FFT, which achieves only 7.8\% top-1 accuracy, leading to a decrease of nearly 68\% from the zero-shot performance. As the flexibility of the adaptation strategy increases, the degradation becomes more pronounced. LoRA reaches 19.9\% and 40.4\% top-1 accuracy on EuroSAT and RESISC45, resulting in decreases of more than 24\% and 28\%, respectively, while maintaining near zero-shot performance on ImageNet. This indicates that preserving zero-shot representations does not necessarily ensure robust transfer across RS datasets.
These results clearly illustrate the severe catastrophic forgetting induced by task-specific federated fine-tuning. This trade-off between task specialization and cross-domain generalization underscores the need for mitigation techniques, as explored in Section \ref{sec:mitigation-techniques}.

\subsection{Mitigating Catastrophic Forgetting}
\label{sec:mitigation-techniques}
% TODO if we keep this we need to remove mentions from the generalization performance part
\begin{figure}[t]
    \centering
    \includegraphics[width=0.49\textwidth]{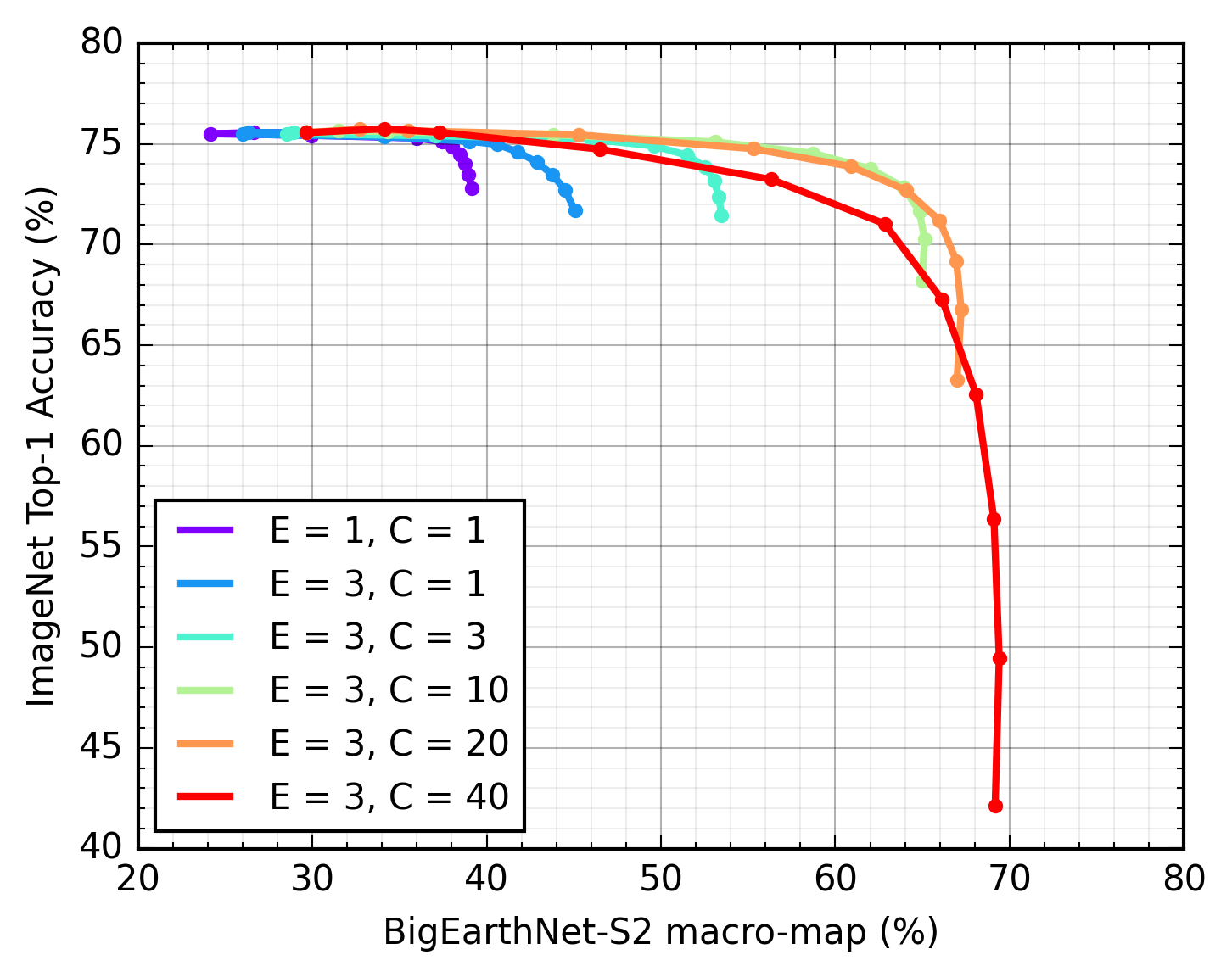} 
    \caption{Trade-off between in-distribution performance on BigEarthNet-S2 and catastrophic forgetting on ImageNet obtained through linear interpolation between zero-shot and federated fine-tuned CLIP weights. Each curve corresponds to a different fine-tuning duration defined by the number of local epochs $E$ and communication rounds $C$. 
    % The right figure shows a comparison of converged models using FFT (CLIP ViT-L/14 and ViT-B/32), and image encoder fine-tuning (ViT-L/14).
    }
    \label{results:tradeoff}
\end{figure}
The results in Section \ref{generalization} indicate severe catastrophic forgetting when fine-tuning CLIP on BigEarthNet-S2, as reflected by the degradation on ImageNet. The effect is most pronounced for FFT, where updating all model parameters leads to substantial degradation performance in natural image classification. This confirms that unrestricted adaptation to the RS task causes the pretrained representations to specialize strongly, thereby reducing their generalization capability. To mitigate this effect, we investigate PAINT \cite{ilharco_patching_2022}, a method designed to reduce catastrophic forgetting on a supported task after training on a new patching task. The central idea of PAINT is to improve performance on the patching task while maintaining performance on the supported task by interpolating between pretrained and fine-tuned weights. After training, both tasks are evaluated across a range of interpolation coefficients to identify an appropriate trade-off point. In the original formulation, this point corresponds to the highest performance on the patching task while allowing at most a 1\% reduction on the supported task. In our setting, we consider ImageNet as the supported task and BigEarthNet-S2 as the patching task. The objective is to increase RS performance while preserving the generalization capabilities of zero-shot CLIP, which may remain relevant for downstream tasks beyond BigEarthNet-S2.

Figure~\ref{results:tradeoff} shows the trade-off between validation performance on BigEarthNet-S2 and ImageNet for different interpolation coefficients and training durations. As one can observe, higher weighting of the fine-tuned weights improves performance on BigEarthNet-S2 while simultaneously degrading performance on ImageNet. In contrast to the original PAINT setting, where training is limited in duration and only the image encoder is fine-tuned, we apply longer training and FFT to a similarly sized CLIP model. Under these conditions, the trade-off becomes less favorable. At convergence, maximizing performance on BigEarthNet-S2 inherently results in significant performance degradation on ImageNet. Extended training is required to reach the maximum performance on BigEarthNet-S2. However, ImageNet accuracy deteriorates rapidly with stronger weighting of the patching task and exceeds the 1\% tolerance threshold even with low weighting of the patching task. For this reason, we additionally evaluate shorter fine-tuning durations. The corresponding curves indicate that intermediate checkpoints provide a more balanced compromise between performance and generalization. Furthermore, after sufficient fine-tuning, interpolating a small portion of pretrained and fine-tuned weights improves performance on both tasks. This indicates that effective interpolation becomes possible only after extended training, whereas shorter training does not sufficiently adapt the representations.

\subsection{Local Training Complexity}
\label{sec:computational_efficiency}

\begin{table*}[htbp]
\renewcommand{\arraystretch}{1.15}
\centering
\caption{Local training complexity and communication cost of the considered adaptation strategies. Local training complexity is reported in terms of FLOPs and MACs per forward pass under training without caching and with caching. Communication cost corresponds to the number of trainable parameters exchanged per communication round in FL.}

\vspace{4mm}
\begin{tabular}{lccccccc}
\toprule
 & \multicolumn{4}{c}{\textbf{Local Training Complexity}} 
 & \multicolumn{2}{c}{\textbf{Communication Cost}} \\
\cmidrule(lr){2-5}
\cmidrule(lr){6-7}
\textbf{Adaptation Strategy} 
& \multicolumn{2}{c}{w/o Caching} 
& \multicolumn{2}{c}{with Caching}
& Total Params 
& Trainable Params \\
\cmidrule(lr){2-3}
\cmidrule(lr){4-5}
& FLOPs & MACs 
& FLOPs & MACs 
&  &  \\
\midrule
Baseline & 2.60B & 1.29B & 2.60B & 1.29B & 23.55M & 23.55M \\
\midrule
Zero-shot & --- & --- & --- & --- & 427.62M & --- \\
FFT & 543.25B & 271.48B & 543.25B & 271.48B & 427.62M & 427.62M \\
Image Encoder Fine-tuning & 543.25B & 271.48B & 155.6B & 77.77B & 427.62M & 303.97M \\
Text Encoder Fine-tuning & 543.25B & 271.48B & 387.66B & 193.72B & 427.62M & 123.65M \\
CoOp & 404.28B & 202.04B & 248.68B & 124.27B & 427.66M & 46.85K \\
LoRA & 545B & 272.36B & 545B & 272.36B & 428.70M & 1.08M \\
\bottomrule
\end{tabular}

\label{results:communication-cost}
\end{table*}

In this subsection, we analyze the local training complexity of the considered approaches. Table 2 reports the required FLOPs and MACs per forward pass under training without caching and with caching. Training without caching corresponds to recomputing image and text features at every iteration, whereas training with caching allows feature reuse for frozen components, thereby reducing computational cost when applicable.

\paragraph{Training Without Caching.}

In the absence of caching, all features must be recomputed in every forward pass. Under this regime, FFT, image-encoder fine-tuning, and text-encoder fine-tuning require the same number of FLOPs and MACs. Although the set of trainable parameters differs, the underlying model architecture and the executed operations remain unchanged, resulting in identical computational complexity per forward pass. CoOp \cite{zhou_learning_2022} exhibits lower FLOPs and MACs during the forward pass compared to the other CLIP-based approaches. While CoOp uses a longer token sequence in the text encoder (maximum context length of 61 tokens compared to 5 learnable context vectors in other approaches), it does not employ prompt ensembling. In contrast, the remaining adaptation strategies construct text features using multiple prompt templates (six in our implementation), which increases the computational cost of the text encoder. As a result, CoOp requires fewer overall operations during feature construction. LoRA \cite{hu_lora_2022} introduces a slight increase in FLOPs and MACs compared to FFT. The additional LoRA modules inserted into the attention layers add extra operations during both the forward and backward passes, leading to moderately higher computational complexity.

\paragraph{Training With Caching.}

When caching is permitted, computational complexity depends on which components of the model remain unchanged during training. FFT does not benefit from caching, as both the image and text encoders are updated. Consequently, image and text features must be recomputed in every iteration, resulting in 543.25B FLOPs and 271.48B MACs per forward pass. In image encoder fine-tuning, the text encoder remains frozen. Therefore, text features can be computed once and cached for subsequent iterations. This reduces the computational cost to 155.6B FLOPs and 77.77B MACs, corresponding to a reduction of approximately 387B FLOPs compared to FFT. The reduction is substantial, particularly because text feature construction is costly due to prompt ensembling. Conversely, in text-encoder fine-tuning, the image encoder remains fixed and image features can be cached. However, text features must be recomputed at each iteration, as the text encoder parameters are updated. Since prompt ensembling is used, this recomputation remains computationally demanding. For CoOp\cite{zhou_learning_2022}, the learnable soft prompts alter the input sequence to the text encoder, causing the resulting text features to change during training. Thus, text features cannot be cached. Nevertheless, the computational cost remains lower than in text encoder fine-tuning, as CoOp does not rely on prompt ensembling despite using longer input sequences. Although LoRA modifies only a small number of parameters, both the image and text encoders are effectively altered through the injected low-rank modules. As a result, image and text features cannot be cached during training. The additional LoRA components further contribute to slightly higher FLOPs and MACs compared to FFT.

\subsection{Communication Cost}
\label{sec:communication_efficiency}

Communication cost is determined by the number of trainable parameters exchanged between clients and the central server at each communication round in FL. Table 2 reports the total and trainable parameters shared with the central server for all considered strategies. Although all CLIP-based models have a similar total parameter count (approximately 427M parameters), the number of trainable parameters varies substantially across adaptation strategies. FFT updates all 427.62M parameters, resulting in the highest communication cost. Freezing either the image encoder or the text encoder reduces the number of communicated parameters to 303.97M and 123.65M, respectively. CoOp achieves the lowest communication cost by a large margin, with only 46.85K trainable parameters. This reduction can be attributed to the optimization and transmission of only a small set of soft prompt vectors. LoRA introduces 1.08M trainable parameters, which is larger than CoOp but still significantly smaller than encoder-specific fine-tuning or FFT. Consequently, LoRA substantially reduces communication overhead while maintaining strong task performance. Although the total parameter count of CLIP exceeds that of the baseline (23.55M parameters), both CoOp and LoRA communicate fewer parameters than the baseline under standard iterative model averaging.

\section{Conclusion}
\label{cha:conclusion}

In this paper, we have presented a comparative study of adaptation strategies for CLIP in the context of FL for RS. Specifically, we investigated FFT, encoder-specific fine-tuning, prompt learning, and LoRA under non-IID data. The considered strategies have been comparatively analyzed with respect to: 
1) in-distribution performance, 
2) out-of-distribution performance, 
3) local training complexity, and 
4) communication cost. Our results demonstrate that adapting a VLM in a federated setting is advantageous over the baseline and zero-shot CLIP. However, the choice of adaptation strategy significantly affects the trade-off between generalization capability, local training complexity, and communication cost. Based on our theoretical and experimental analyses, we derive the following guideline for selecting suitable adaptation strategies in federated VLM training for RS:

\begin{enumerate}
    \item \textbf{Text encoder fine-tuning} offers the strongest cross-dataset generalization among the fine-tuned models and exhibits moderate communication cost. It represents a balanced choice when the transfer to related RS datasets is a primary objective.
    
    \item \textbf{Prompt Learning} minimizes communication cost, requiring only a negligible fraction of trainable parameters. However, it achieves lower task accuracy and weaker cross-dataset generalization than most of the other adaptation strategies considered in this work. This strategy should be considered only when communication bandwidth is the dominant constraint.

    \item \textbf{LoRA} provides the most favorable trade-off between performance and communication efficiency. It achieves competitive in-distribution performance while reducing the number of communicated parameters by more than two orders of magnitude compared to FFT. It is therefore recommended when both performance and communication constraints are important.
    
    \item \textbf{FFT combined with PAINT} enables controlled adaptation by interpolating pretrained and fine-tuned weights. Although it inherits the high communication cost of FFT, it demonstrates the potential to extend a zero-shot model to a new task while largely preserving performance on a previously learned task. 
\end{enumerate}
Overall, our findings indicate that parameter-efficient adaptation strategies are particularly well suited for federated VLM training in RS, as they mitigate communication cost while maintaining competitive performance. At the same time, unrestricted FFT leads to substantial catastrophic forgetting, limiting its applicability in settings where generalization beyond the target task is required. 

As a future work, we plan to extend the analysis to more advanced state-of-the-art federated VLM adaptation strategies. Furthermore, we aim to extend the analysis to domain-specific foundation models to determine if RS-specific inductive biases alter the comparative trade-offs observed with general-purpose VLMs. In addition, we aim to expand this framework beyond RGB imagery to evaluate these strategies on multispectral bands and settings that integrate different sensor modalities.

%Furthermore, we do not systematically vary the degree of statistical heterogeneity across clients. Consequently, the robustness of the considered adaptation strategies under varying non-IID conditions remains to be investigated.

% References
\bibliographystyle{spiebib} % makes bibtex use spiebib.bst
\bibliography{main} % bibliography data in report.bib

\end{document}